\documentclass[10pt,twocolumn,letterpaper]{article}

\usepackage{cvpr}
\usepackage{times}
\usepackage{epsfig}
\usepackage{graphicx}
\usepackage{amsmath}
\usepackage{amssymb}
\usepackage[font={footnotesize},justification=centering]{subcaption}
\usepackage[font=small,labelfont=bf]{caption}
\usepackage{physics}
\usepackage{array}
\usepackage{multirow}
\usepackage{scrextend}
\usepackage{cite}
\usepackage{verbatim}
\usepackage{indentfirst}

\graphicspath{ {images/} }
\usepackage[breaklinks=true,bookmarks=false]{hyperref}
\cvprfinalcopy

\begin{document}

\title{AI Blue Book: Vehicle Price Prediction using Visual Features}

\author{Richard R. Yang \hspace{0.5cm} Steven Chen \hspace{0.5cm} Edward Chou \hspace{0.5cm} \\
Department of Computer Science\\
Stanford University\\
\{richard.yang, stevenzc, ejchou\}@stanford.edu
}

\maketitle


\begin{abstract}
In this work, we build a series of machine learning models to predict the price of a product given its image, and visualize the features that result in higher or lower price predictions. We collect two novel datasets of product images and their MSRP prices for this purpose: a bicycle dataset and a car dataset. We set baselines for price regression using linear regression on histogram of oriented gradients (HOG) and convolutional neural network (CNN) features, and a baseline for price segment classification using a multiclass SVM. For our main models, we train several deep CNNs using both transfer learning and our own architectures, for both regression and classification. We achieve strong results on both datasets, with deep CNNs significantly outperforming other models in a variety of metrics. Finally, we use several recently-developed methods to visualize the image features that result in higher or lower prices.
\end{abstract}

\section{Introduction}

Online shopping is quickly becoming the norm, but the experience differs greatly from retail shopping, in which people have the opportunity to closely examine a product, weighing in the feel of a material or the scent of a cream before making a purchase decision. Online shoppers must rely entirely on the few product images to make a decision.

In this work, we build, optimize, and evaluate an ensemble of machine learning models that can predict prices based on product images, for both regression and classification tasks. These models can be used by both buyers and sellers to suggest fair prices for products, or warn of inaccurate or unreasonable pricing. In this work, we also visualize which features tend to result in predicted higher or lower prices. Our proposed model can help sellers increase the perceived value of their products, helping guide product design and photo selection to improve a buyer's impression.

\section{Related Work}

Computer vision and supervised machine learning have been used in conjunction for a variety of pricing and regression tasks. Early work has used supervised learning to predict attractiveness given labeled faces~\cite{attractiveness}. Recent work have predicted age using face images~\cite{ageregression,ageestimation}, and housing prices with satellite imagery~\cite{housepriceprediction,imagerealestate}, tasks which are traditionally difficult for humans to perform accurately. In contrast to these prior work, we focus on the task of prediction using images of consumer products, with novel datasets curated specifically for this purpose.

ClickToPrice~\cite{clicktoprice} proposes the most similar concept to our work. In ClickToPrice, the author explores the predictive power of product images for prices. Our project is similar in that we use machine learning to predict the prices. However, ~\cite{clicktoprice} uses basic techniques to perform classification into a the general product category (e.g., towels, shoes), and uses that categorization alone to predict the category average price for each item. We argue that such a model is functionally equivalent to image classification, and is not suited for price prediction. Our models are specifically designed for fine-grained price prediction for items of the same type and are significantly more sophisticated in technical implementation and more accurate on individual image queries.

Recent research has delved into methods for visualizing what features and image parts CNNs use to determine their predictions. Zeiler and Fergus~\cite{visualizingnetworks} learn what visual features maximize hidden unit activations, and use obscuring sliding windows to determine which features influence prediction. Yosinski et al.~\cite{deepvisualization} build live visualizations of activations, allowing for easier discovery of the inner workings of CNNs. Zhou et al.~\cite{learningdeepfeatures} use global average pooling to visualize what regions of images are most responsible for classification predictions. In contrast, Simonyan et al.~\cite{deepinsidenetworks} generate images that maximize the class score predicted by an object recognition network. We experiment with a subset of these methods to visualize the features that result in higher or lower prices for products.

\section{Approach}

We first present the datasets we collect specifically for this work, then describe the algorithmic models used to predict prices within these datasets.

\subsection{Datasets}

In this work, we choose to use bicycles and cars as target product datasets, due to the wide visual variances in bike and car models, close visual correlations to prices, and relevance of online shopping for cars and bikes.

\begin{figure}[t]
    \centering
    \includegraphics[width=0.52\linewidth]{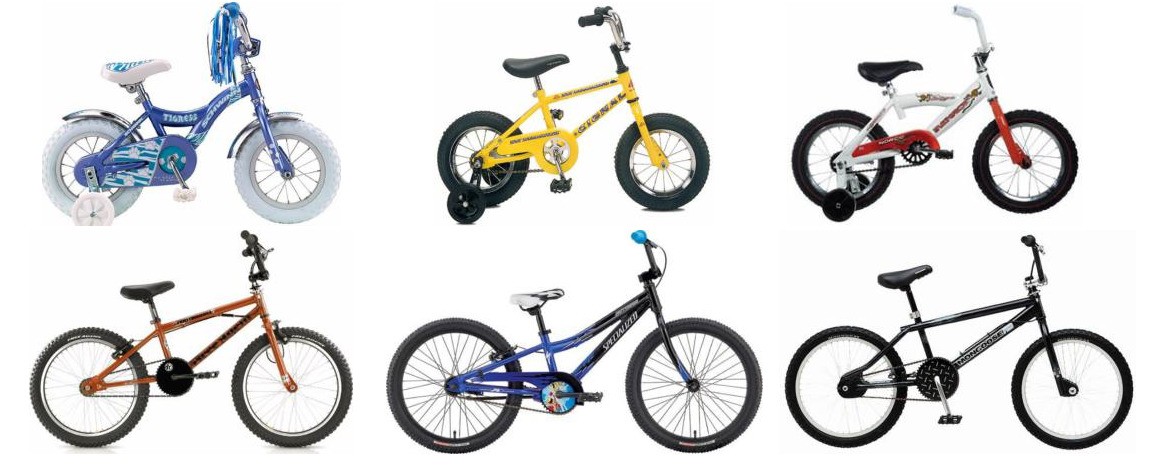}
    \hfill
    \includegraphics[width=0.45\linewidth]{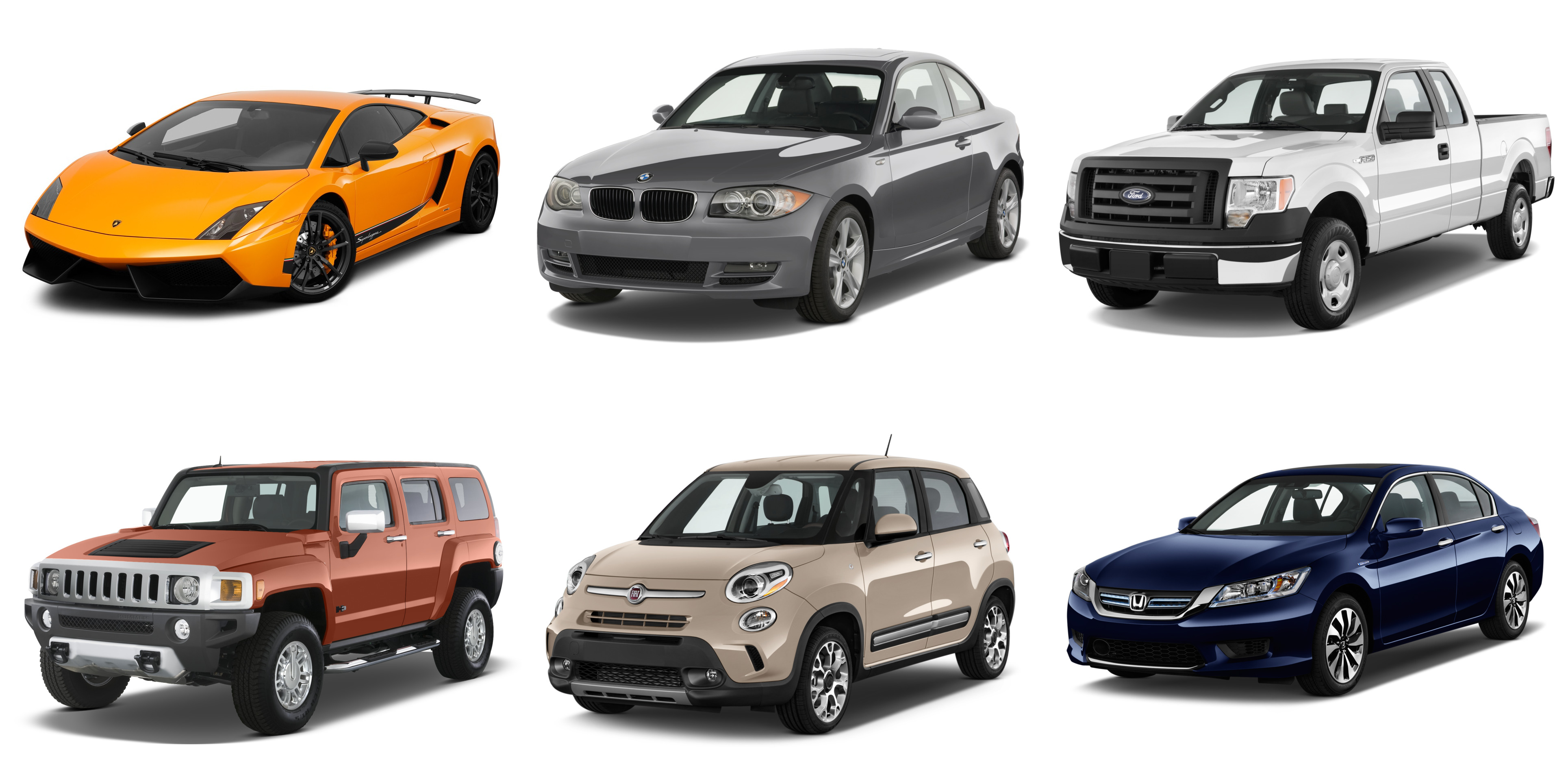}
    \centering
    \caption{Examples of images from bikes (left) and cars(right).}
    \label{fig:bike_dataset}
\end{figure}

Our first dataset, \textbf{bikes}, is curated from an online database for bicycle valuation. We collect images and prices from the listings, and preprocess by filtering out low quality images and resizing to 224 by 224 pixels. Our final dataset contains solid background, side view images. The dataset consists of 21,843 images, each labeled with an MSRP (manufacturer's suggested retail price).

Our second dataset, \textbf{cars}, is a dataset of vehicle images and their prices. We retrieve price data from Kaggle\footnote{\texttt{www.kaggle.com/jshih7/car-price-prediction}}. We join these prices on images from Google Images, using search terms consisting of model and year, along with ``Angular Front View". We clean and resize the images, resulting in a final dataset of 1,400 examples.

\begin{figure}[t]
    \centering
    \includegraphics[width=0.48\linewidth]{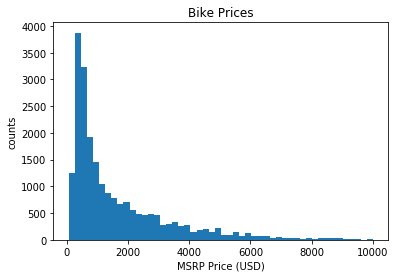}
    \hfill
    \includegraphics[width=0.48\linewidth]{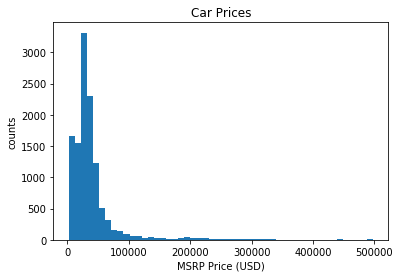}
    \caption{Histogram of prices for bikes (left) and cars (right).}
    \label{fig:bike_cdf}
\end{figure}

The bicycle prices range between \$70 and \$17,000, and the car prices range between \$12,000 and \$2,000,000 (see Figure \ref{fig:bike_cdf}). The prices closely follow an exponential CDF distribution, in which there are significantly more models at the low and regular price segments than at the luxury segment.

\subsection{Modeling}

We approach price prediction through two different learning objectives: regression and classification. In the regression models, we attempt to directly predict the numerical price given an image. In the price classification models, we split our data into various price segments and treat it as classification into price ranges.

\subsubsection{Linear Regression Baselines}

Our first baseline is multiple (multivariable) linear regression using histogram of oriented gradients (HOG) features, using PCA to reduce overfitting. Our second regression model is multiple linear regression using CNN features. For this model, we generate CNN features from the last convolutional layer of VGG-16~\cite{vgg}, a CNN pre-trained on ImageNet for object recognition, then use PCA-reduced-dimension features as input data. We report parameter values and evaluate performance for this baseline and the others in Section \ref{results}. 

\subsubsection{Multiclass SVM Baseline}

Our baseline for classification is a multiclass linear support vector machine (SVM) trained on price segments (see Section \ref{results} for segments). To support multiclass classification, we use the one vs. one approach, which trains one binary SVM between each pairwise combination of categories. Each binary classifier votes for a category, and the prediction of the model is the category that received the most votes.

\subsubsection{Transfer Learning CNN}

Our first CNN models are trained using transfer learning. In particular, we use the pre-trained ImageNet image recognition networks VGG-16 and SqueezeNet. VGG-16~\cite{vgg} is a large CNN architecture consisting of many layers of small convolution and pooling filters, followed by two fully connected layers and a softmax output, with a total of 138M parameters. SqueezeNet~\cite{iandola2016squeezenet} is a recent CNN that achieves AlexNet~\cite{alexnet} level performance while only having 1.3M parameters. For both models, we use the Keras~\cite{keras} framework, load the architecture and weights, and remove the networks' softmax and dense layers. We set the remaining layers to be fixed, and add our own fully connected layer.

We use two different output layers, each designed for a specific task. For continuous price regression, we add a single linear activation output unit after the fully connected layer. For segmented price classification, we add an output layer with an output unit for each class, and use a softmax activation. We optimize and tune each network and task pair separately (see Section \ref{results} for more details).

\subsubsection{PriceNet}

\begin{figure*}
  \includegraphics[width=\textwidth]{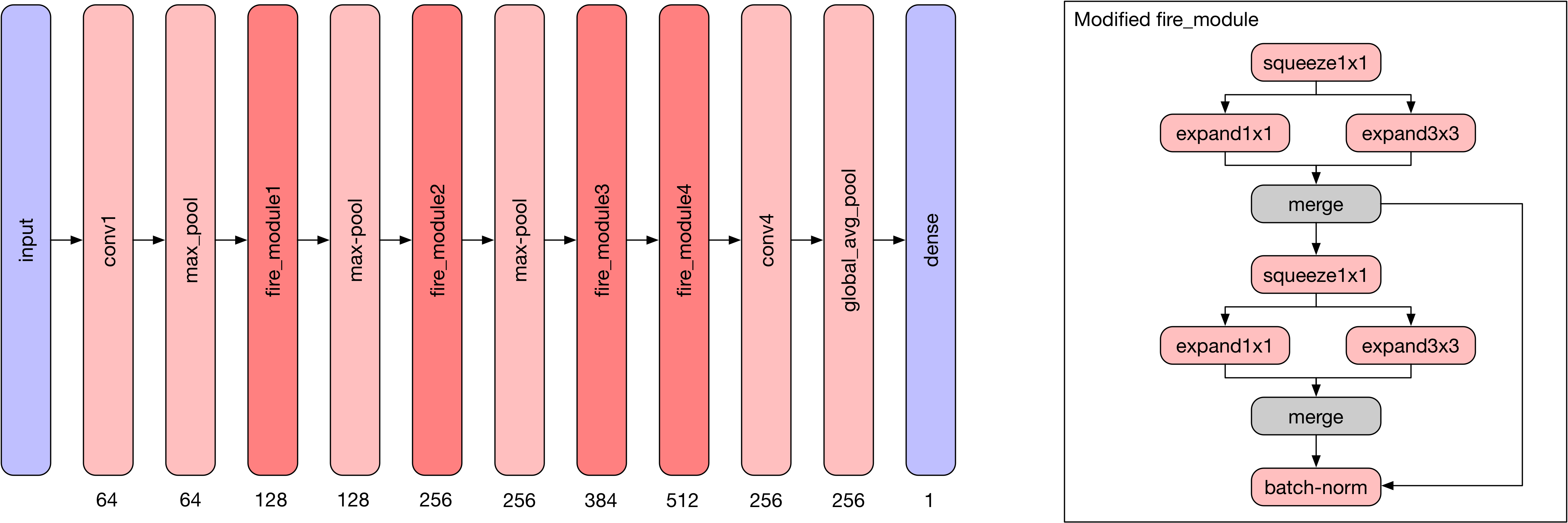}
  \caption{Left: The full PriceNet architecture. The values at the bottom denote the output depth of each layer. Right: Our modified fire module, with residual connections and batch normalization.}
  \label{pricenet}
\end{figure*}

Lastly, we design our own deep learning architecture called PriceNet, which is an expansion on the SqueezeNet architecture. SqueezeNet has a small number of parameters in comparison to other model architectures because expensive 3x3 convolutions are replaced with \textit{fire modules}. In a fire module, the depth of the volume is first downsampled by efficient 1x1 convolutions (squeeze), then upsampled by a combination of 1x1 and 3x3 convolutions (expand) ~\cite{iandola2016squeezenet}. We modify the SqueezeNet architecture by introducing residual connections between the fire layers, and adding batch normalization to each fire module. The full PriceNet architecture is shown in Figure \ref{pricenet}. We build two variations of PriceNet: PriceNet-Reg with a linear output activation for price regression, and PriceNet-Class with a softmax output activation for price segment classification. Both networks contain around 1.2M parameters. We tune the two networks separately.

\section{Experimental Results} \label{results}

We first describe our tuning process and parameter selections for our models. We then present our evaluations of our models in terms of several metrics. Finally, we present several different visualizations from our CNNs, along with our interpretations.

\subsection{Model Tuning and Parameters}

\subsubsection{Linear Regression Baselines}

For linear regression with HOG, we generate HOG features with 8 orientations per histogram and a window size of 32 by 32 pixels, selected as a reasonable balance between resolution and noise. We then run principal components analysis (PCA) and reduce the feature dimensionality to 200. For linear regression with CNN features, we generate features using the last convolutional layer of VGG-16~\cite{vgg}, and use PCA to reduce the feature dimensionality to 256.

\subsubsection{Multiclass SVM Baseline}

We tune the SVM model with respect to two hyperparameters: $C$ and $\gamma$, and run hyperparameter search on a log scale of the parameters. The top performance converges after using $C \geq 1$ with a wide range of $\gamma$, so we select $C=1$ and $\gamma = 0.001$.

\subsubsection{Transfer Learning}

While training our transfer learning models, we use several techniques to tune our weights and parameters. We first augment our training data by applying the following transformations randomly in a batch: crop, flip, scale, translate, rotate and Gaussian blur. 

For both regression and classification, we obtain the highest performance with the RMSprop optimizer, which divides learning rate by an exponentially decaying average of squared gradients. We also use dropout, dropping the effects of random hidden units during training, to help reduce overfitting. For parameter selection, we first tune parameters over a log scale and find the best candidates, and then fine tune over a smaller range around the candidates. We tune the learning rate, minibatch size, number of hidden units, and number of training epochs.

\subsubsection{PriceNet}
We train PriceNet-Reg and PriceNet-Class from scratch on the bikes dataset. Due to of the small size of our car dataset (1,400 car images compared to more than 20,000 bike images), we did not have enough data to train a deep neural network from random weight initialization for cars. As a result, we initialize the network with SqueezeNet weights trained on ImageNet and use Glorot initialization for the remaining layers. During training, we perform the same image augmentation techniques as for our transfer learning models. We tune both PriceNet-Reg and PriceNet-Class using log-scale parameter selection to tune learning rate, minibatch size, number of hidden units, and number of training epochs.

\subsection{Evaluation}

We split our datasets into training and testing splits, which are consistent across all models to ensure fair comparison. To create the split for both datasets, we first shuffle, then assign 90 percent of the points to train, and the remaining 10 percent to test.

\subsubsection{Regression Models}

We use three different metrics to evaluate and compare the performance of our models on price regression: root mean squared error (RMSE), mean absolute error (MAE), and coefficient of determination ($R^2$). RMSE measures the root average squared error between the predicted and actual price $\sqrt{\sum_{i=1}^m (y^{(i)} - \hat{y}^{(i)})^2}$, while MAE is interpreted as the average absolute difference in price $\sum_{i=1}^m \abs{y^{(i)} - \hat{y}^{(i)}}$. Lower values are better for both. Coefficient of determination measures the proportion of variance explained by the model, and lies between 0 and 1, where higher values are better.

We report results of our two linear regression models and the deep CNN in Table \ref{tab:bikesregression} (bikes) and \ref{tab:carsregression} (cars), alongside a naive baseline that always predicts the average price. All models significantly outperform the naive baseline, with linear regression on CNN features showing a margin of improvement over HOG, likely due to the CNN features providing more discriminative visual cues. The deep CNNs in particular have very strong performance, significantly outperforming the other models in every metric. On the bikes dataset, our PriceNet architecture achieves the strongest results in each metric, with an MAE of \$165.87 on prices ranging from \$70 to \$1,700. On the cars dataset, the SqueezeNet transfer CNN and PriceNet achieves similar performance.

\begin{table}
    \centering
    \begin{tabular}{l | r r r}
    \textbf{Model}                 & \textbf{RMSE} & \textbf{MAE} &\textbf{ $R^2$} \\
    \hline
    Average Baseline      & 1810.19 & 1318.53 & 0.00  \\
    \hline
    LinReg (HOG Features) & 1274.99 & 833.02 & 0.50  \\
    \hline
    LinReg (CNN Features) & 1054.67 & 712.63 & 0.66  \\
    \hline
    VGG16 Transfer     & 747.42 & 405.50 & 0.83  \\
    \hline
    SqueezeNet Transfer      & 720.19 & 403.38 & 0.84  \\
    \hline
    \textbf{PriceNetReg}     & \textbf{262.78} & \textbf{165.8}7 & \textbf{0.98} \\
    \end{tabular}
    \caption{Test set results for regression on the bikes dataset.}
    \label{tab:bikesregression}
\end{table}

\begin{table}
    \begin{tabular}{l | r r r}
    \textbf{Model}                 & \textbf{RMSE} & \textbf{MAE} & \textbf{$R^2$} \\
    \hline
    Average Baseline      & 76240.41 & 44410.57 & 0.00  \\
    \hline
    LinReg (HOG Features) & 41898.48 & 27588.70 & 0.70  \\
    \hline
    LinReg (CNN Features) & 37808.84 & 23929.67 & 0.75  \\
    \hline
    VGG16 Transfer      & 12363.65 & 7477.74 & 0.97  \\
    \hline
    \textbf{SqueezeNet Transfer }     & \textbf{10577.47} & 6953.01 &\textbf{ 0.98}  \\
    \hline
    \textbf{PriceNetReg} & 11587.05 & \textbf{5051.61} & \textbf{0.98}  \\
    
    \end{tabular}
    \caption{Test set results for regression on the cars dataset.}
    \label{tab:carsregression}
\end{table}

\subsubsection{Classification Models}

For classification, we assign class segments to each example using price cutoffs corresponding to percentiles of price. We assign labels of 25, 50, 75, 100 for the bikes dataset (4 classes), and 20, 40, 60, 80, 100 for the cars dataset (5 classes). While classification does not predict price directly like regression, we have two main reasons for using classification; price segmentation is useful in many business applications, and classification allows us to apply certain techniques such as class activation maps to visualize features. We evaluate our classification models on three primary metrics: precision, recall, and the F-1 score.

\begin{table}[]
\centering
\begin{tabular}{l|lll}
\textbf{Model}    & \textbf{Precision} & \textbf{Recall} & \textbf{F1-Score} \\ \hline
SVM (bikes)      & 0.80               & 0.45            & 0.43              \\ \hline
VGG16 Transfer (bikes) & 0.74               & 0.75            & 0.74              \\ \hline
\textbf{PriceNetClass (bikes)} & \textbf{0.89}               & \textbf{0.88}            & \textbf{0.88}              \\ \hline
\hline
SVM (cars)       & 0.83               & 0.82            & 0.82              \\ \hline
VGG16 Transfer (cars)       & 0.82               & 0.82            & 0.82              \\ \hline
\textbf{PriceNetClass (cars)}  & \textbf{0.88}               & \textbf{0.88}            & \textbf{0.88}             
\end{tabular}
\caption{Classification results for the bikes and cars datasets.}
\label{classification-summary}
\end{table}

\subsection{Visualizations} \label{visualizations}

In this section, we describe the three different methods we use to visualize how our deep CNN models see their input and which visual features affect the models' perception.

\subsubsection{Sliding Window Heatmaps}

We use obscuring sliding windows on a deep CNN with linear output (regression) to determine which features of input images are important to determining the predicted price, in the vein of~\cite{visualizingnetworks}. We slide a 28 by 28 pixel window over an input image, obscuring that area of the image by replacing the pixel values with the average value for the network. We then run the obscured images through the network, and compare the predicted price of the obscured images to the original predicted price.

\begin{figure}[t]
    \centering
    \includegraphics[width=0.7\linewidth]{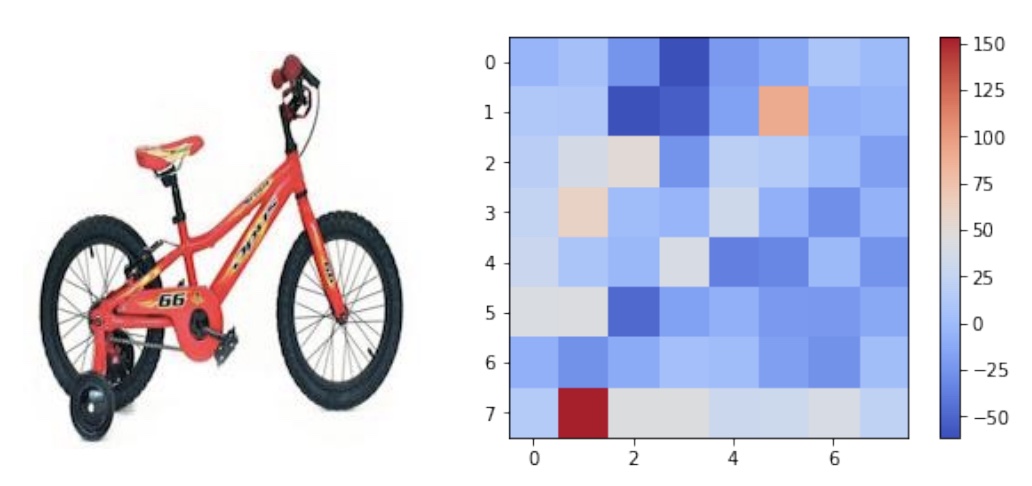}
    \caption{Sliding window heatmap visualizing obscuring regions that result in higher (red) or lower (blue) prices. Obscuring the training wheels (bottom left), a visual cue of a low-price model, results in the predicted price increasing by 150 dollars.}
    \label{fig:kidbike}
\end{figure}

We visualize these changes using a heatmap, where each square of the heatmap corresponds to the region of the image that was obscured. An example is shown in Figure \ref{fig:kidbike}: obscuring the training wheels increases the predicted price by \$150.

\subsubsection{Saliency Maps}

\begin{figure}[t]
    \centering
    \includegraphics[width=0.7\linewidth]{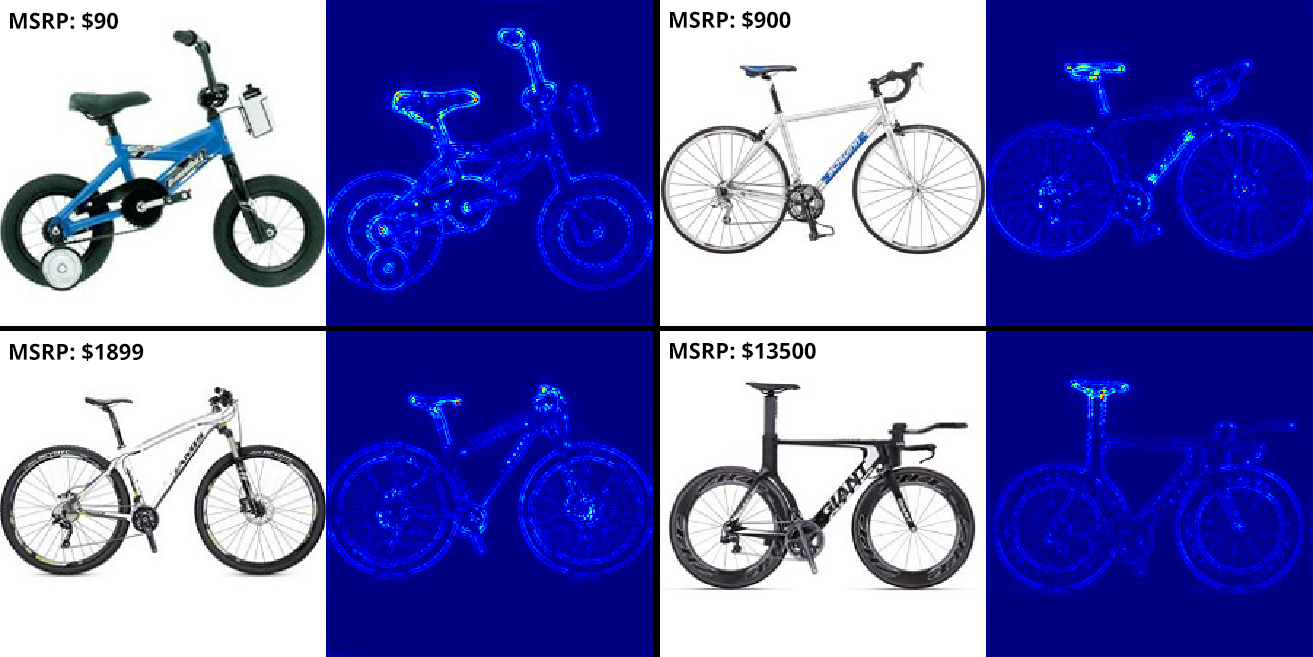}
    \caption{A comparison of saliency maps for bike examples across four different classes. Note the highlights on the seat, handlebars, gearbox, and wheel spokes.}
    \label{fig:bike_saliency_map}
\end{figure}

\begin{figure}[t]
    \centering
    \includegraphics[width=0.7\linewidth]{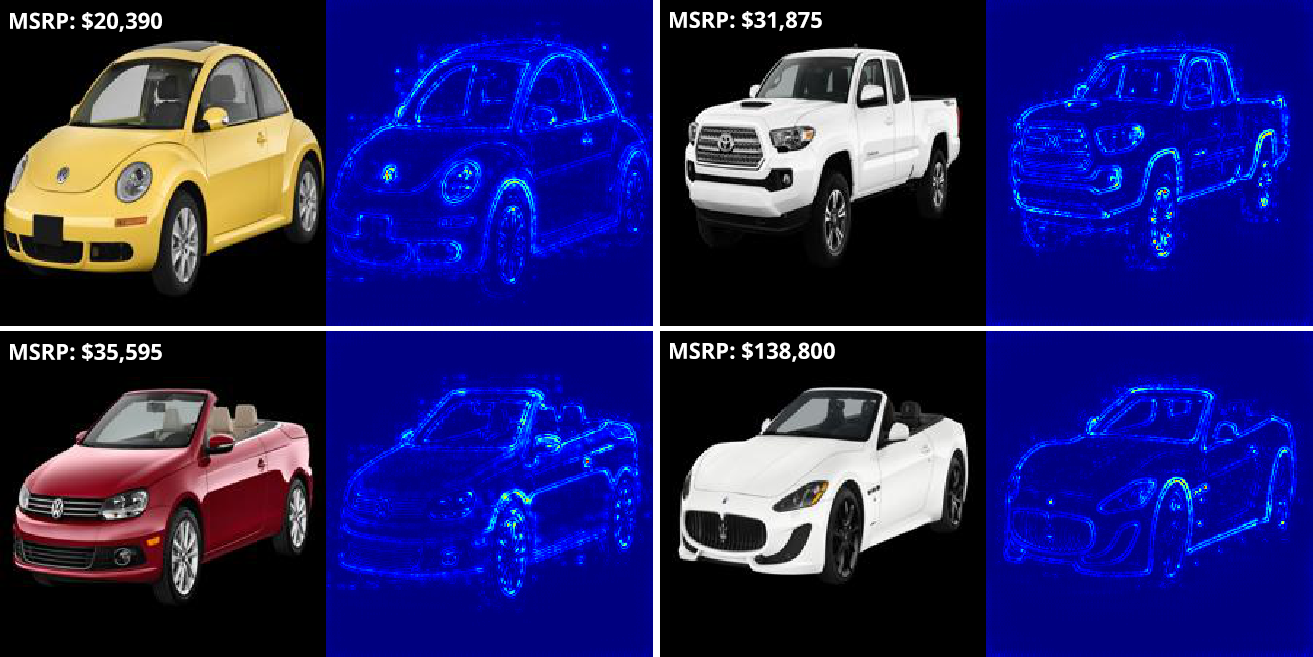}
    \caption{A comparison of saliency maps for car examples across four different classes. Note the highlights on the manufacturer logo, frame shape, and tires.}
    \label{fig:car_saliency_map}
\end{figure}

For our classification networks, we create saliency maps to visualize how individual pixels contribute to the output. In \cite{deepinsidenetworks}, the authors demonstrated that the class score $S_C$ of an image $I$ can be approximated by the first-order Taylor expansion $S_C(I) = w^TI+b$, where $w$ is calculated by taking the gradient of the prediction with respect to an input image. $w$ corresponds to the weight of each pixel in $I$, and shows much each pixel contributes to the class prediction.

In Figure ~\ref{fig:bike_saliency_map}, we show the original input images and the respective saliency map for bike examples across the four output classes. From these visualizations, we observe the most salient regions to our model are the seat shape, handlebars, gearbox, and brakes. Similarly, saliency maps for cars are shown in Figure ~\ref{fig:car_saliency_map}. For cars, the most salient regions are the logo, body contour, and wheels.

\begin{figure}[t]
    \centering
    \includegraphics[width=0.49\linewidth]{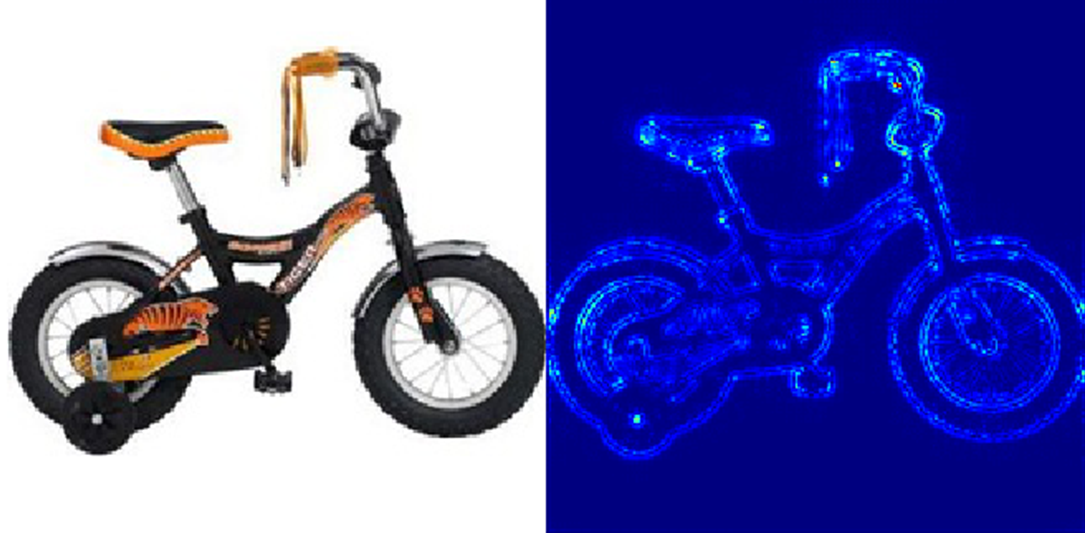}
    \hfill
    \includegraphics[width=0.49\linewidth]{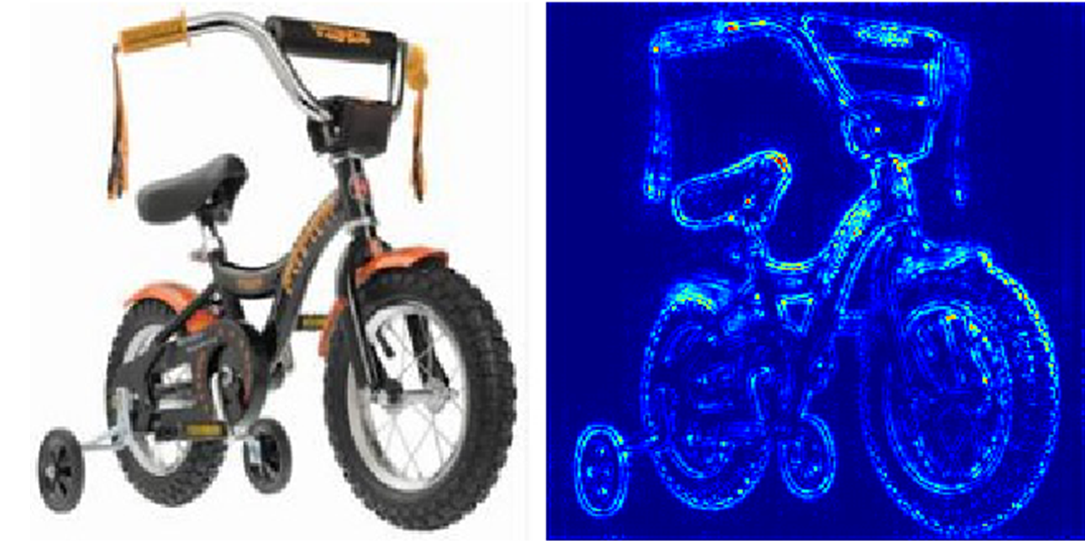}
    \caption{Saliency maps for two perspectives of the same bike.}
    \label{fig:saliency_bike_angles}
\end{figure}

Additionally, we use saliency maps to observe how well our CNNs generalize. We show that our model is invariant to angles by passing two photos of the same bike taken at different angles to our model, and it predicts the correct class for both images. (see Figure \ref{fig:saliency_bike_angles}).

\subsubsection{Gradient-Weighted Class Activation Maps}

\begin{figure}[t]
    \centering
    \includegraphics[width=0.40\linewidth]{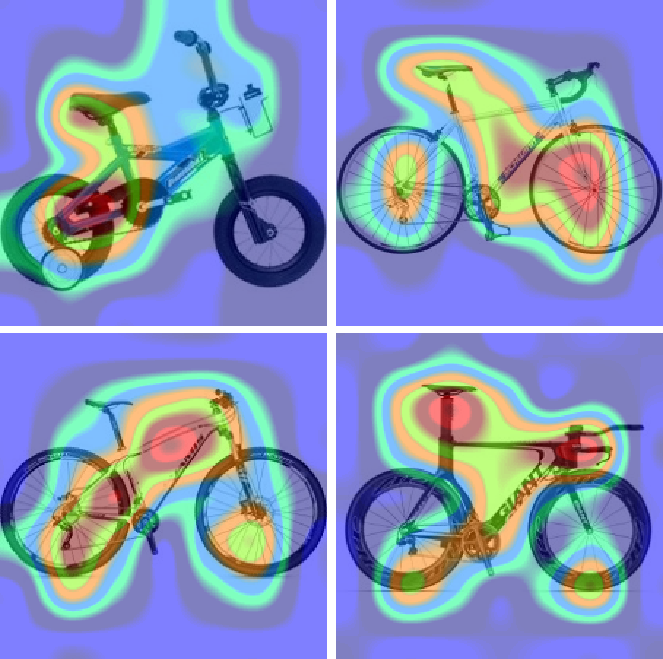}
    \hfill
    \includegraphics[width=0.40\linewidth]{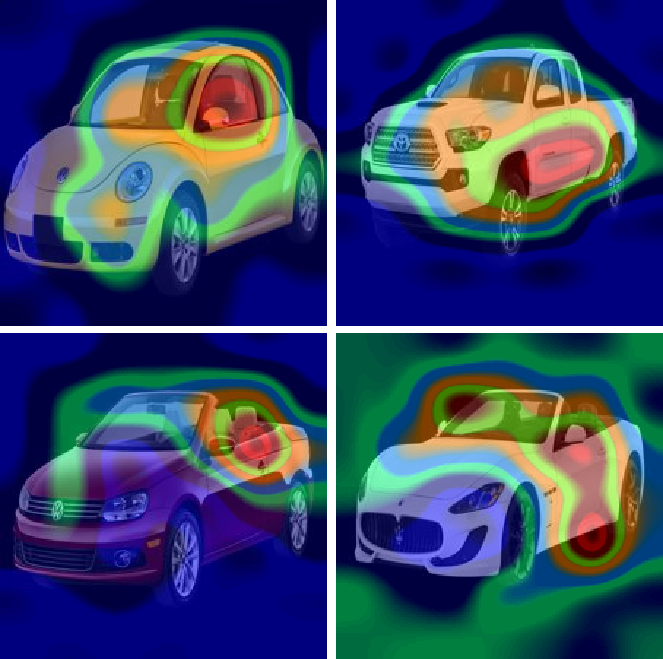}
    \caption{CAM visualizations overlaid on the original input image. Notable areas for bikes include seats, training wheels, and tire spokes. For cars, notable areas include convertible top, doors, and body contours.}
    \label{fig:bike_cam}
\end{figure}

Gradient-weighted Class Activation Maps (CAM) have been previously used for object locality detection~\cite{gradcam}. When multiple output classes are present in an image, CAM will highlight parts of the image that contribute most to the selected output class. Since our classification segments by price, we use CAM to highlight areas of the bike or car that result in cheap or expensive price ranges.

The intuition behind the CAM approach is similar to that for the saliency maps method. However, rather than computing the gradient with respect to the output, we compute the gradient with respect to the feature maps generated by a specific convolutional layer. Then, we apply global average pooling on the gradients to create a weight vector $w$ representative of the contributions of each unit on the class output. In Figure \ref{fig:bike_cam}, we show CAM heatmaps overlaid on the input images. Our model focuses on visually diverse areas of bikes, such as the handlebars, seat, and tires, as well as important regions of cars, such as the convertible top or doors.

\section{Conclusion and Future Work}

In this work, we introduce two novel datasets and build multiple models for predicting prices of products using only single image data. For regression, our custom network architecture PriceNet significantly outperforms multiple transfer learning as well as linear regression baselines. For classification, our transfer learning deep network significantly outperforms models for classifying into price categories. Additionally, we visualize what image regions the deep CNN models discriminate price with using three different methods, providing insight on which visual features of products result in certain prices. 

We have identified multiple future real-world applications of this work. Using feature visualization, merchants can determine what features of objects are correlated to higher prices, and use this to help suggest and guide product design. Our models can also be extended to assist valuations at a large scale, such as for used car sales, where many products must be appraised quickly and prices can be difficult to determine. Finally, our model can be applied to auction sites such as eBay for recommendations for starting bids, and be provided a tool to storefronts and individuals for choosing better photos with higher predicted valuations when listing products for sale.

\newpage

\bibliographystyle{ieee}
\bibliography{egbib}

\begin{thebibliography}{10}\itemsep=-1pt

\bibitem{keras}
F.~Chollet et~al.
\newblock Keras.
\newblock \url{https://github.com/fchollet/keras}, 2015.

\bibitem{attractiveness}
Y.~Eisenthal.
\newblock {\em Facial Attractiveness: Beauty and the Machine}.
\newblock PhD thesis, Tel-Aviv University, 2006.

\bibitem{ageregression}
Y.~Fu and T.~Huang.
\newblock Human age estimation with regression on discriminative aging
  manifold.
\newblock {\em IEEE Transactions on Multimedia}, 10(4), 2008.

\bibitem{ageestimation}
H.~Han, C.~Otto, and A.~Jain.
\newblock Age estimation from face images: Human vs. machine performance.
\newblock In {\em IAPR International Conference on Biometrics}, 2013.

\bibitem{iandola2016squeezenet}
F.~N. Iandola, S.~Han, M.~W. Moskewicz, K.~Ashraf, W.~J. Dally, and K.~Keutzer.
\newblock Squeezenet: Alexnet-level accuracy with 50x fewer parameters and< 0.5
  mb model size.
\newblock {\em arXiv preprint arXiv:1602.07360}, 2016.

\bibitem{alexnet}
A.~Krizhevsky, I.~Sutskever, and G.~E. Hinton.
\newblock Imagenet classification with deep convolutional neural networks.
\newblock In F.~Pereira, C.~J.~C. Burges, L.~Bottou, and K.~Q. Weinberger,
  editors, {\em Advances in Neural Information Processing Systems 25}, pages
  1097--1105. Curran Associates, Inc., 2012.

\bibitem{housepriceprediction}
V.~Limsombunchai.
\newblock House price prediction: Hedonic price model vs. artificial neural
  network.
\newblock In {\em NZARES}, 2004.

\bibitem{clicktoprice}
A.~Maurya.
\newblock Clicktoprice: Incorporating visual features of product images in
  price prediction.
\newblock In {\em INFORMS}, 2016.

\bibitem{gradcam}
R.~Selvaraju, M.~Cogswell, A.~Das, R.~Vedantam, D.~Parikh, and D.~Batra.
\newblock Grad-cam: Visual explanations from deep networks via gradient-based
  localization.
\newblock In {\em ICCV}, 2017.

\bibitem{deepinsidenetworks}
K.~Simonyan, A.~Vedaldi, and A.~Zisserman.
\newblock Deep inside convolutional networks: Visualising image classification
  models and saliency maps.
\newblock In {\em ICLR}, 2014.

\bibitem{vgg}
K.~Simonyan and A.~Zisserman.
\newblock Very deep convolutional networks for large-scale image recognition.
\newblock In {\em ICLR}, 2015.

\bibitem{deepvisualization}
J.~Yosinski, J.~Clune, A.~Nguyen, T.~Fuchs, and H.~Lipson.
\newblock Understanding neural networks through deep visualization.
\newblock In {\em IMCL}, 2015.

\bibitem{imagerealestate}
Q.~You, R.~Pang, L.~Cao, and J.~Luo.
\newblock Image-based appraisal of real estate properties.
\newblock {\em IEEE Transactions on Multimedia}, 19(12), 2017.

\bibitem{visualizingnetworks}
M.~Zeiler and R.~Fergus.
\newblock Visualizing and understanding convolutional networks.
\newblock In {\em ECCV}, 2014.

\bibitem{learningdeepfeatures}
B.~Zhou, A.~Khosla, A.~Lapedriza, A.~Oliva, and A.~Torralba.
\newblock Learning deep features for discriminative localization.
\newblock In {\em CVPR}, 2016.

\end{thebibliography}

\end{document}